# Bi-stable Hair Clip Mechanism for Faster Swimming Robots

Zechen Xiong[1, *], Liqi Chen[2], Hod Lipson[2]

[1]Department of Earth and Environmental Engineering, Columbia University, New York, NY 10027, USA
[2]Department of Mechanical Engineering, Columbia University, New York, NY 10027, USA
[*]Correspondence: zechen.xiong@columbia.edu

## Abstract

Inspired by the snap-through action of a hair clip, we propose to use a bi-stable hair-clip-like mechanism to increase the locomotion ability of soft/compliant robotics. Corresponding mathematical models and analytical theory are built to help us understand and design these robots and manipulators (*1*, *2*). We perform finite element simulation and build hair clip mechanism (HCM) samples to demonstrate the advantages and verify the analytical results. The HCMs are mounted on two different soft fish robots. The pneumatic HCM fish swim at 26.54 cm/s or 1.40BL/s, about twice as fast as a traditional counterpart. The motor-driven HCM fish has a speed of 2.03 BL/s or 42.6 cm/s, 2-3 times faster than previous untethered soft robotic fish. We hope to find a solution to the speed challenge of soft and compliant robots and shed light on novel aquatic propulsion methods.

**Summary:** designing hair-clip-like in-plane prestressed bi-stable mechanisms and using them on soft/compliant fish robots for better performance

## Main Text

### Introduction

Bistable mechanisms are found in nature and technology to perform rapid and repetitive tasks like animal hunting, organism/robotic locomoting, and fast deformation (*3–5*). These mechanisms achieve large shape transitions in a short amount of time while being elastic and reversible. On the other hand, prestressing elevates an elastic structure's energy level and increases the energy-releasing rate when the structure is triggered. Examples of instability and prestressing are common. Many plants can build up elastic energy through osmosis and turgor pressure. A sudden release of this energy can cause rapid movements and plays a critical role in their functions like reproduction and nutrition: the ballistic seed dispersal of Impatiens and squirting cucumber after prestressing their seedpods (*6*); the pollen dispersal of trigger plants (*7*); the rapid leave closure of the iconic Venus flytrap (*8*), etc. However, the build-up of turgor pressure takes large amounts of time and thus cannot support the continuous operation of the prestressed mechanisms.

Interestingly, this prestress-and-snap mechanism is comparatively less observed in the animal kingdom. Just like animals do not evolve wheels, they cannot grow prestressed or buckling mechanisms due to biological constraints. Prestressing and instability lead to additional stress and strain on animals' motor systems, which can lead to fatal damage. The closest examples, though, would be bird feet that use energy-storing tendons to sleep on a branch (*9*) and hummingbird beaks that close fast to capture insects in the air (*3*).

Compared to traditional robotic mechanisms made of rigid links and joints, soft mechanisms are safe, versatile, and bio-compatible because of their elastic and compliant materials. Still, they are intrinsically weaker in force exertion and fast-moving (*10*, *11*). The moduli of soft materials are usually in the order of $10^4$-$10^9$ Pa (*12*). In many cases of soft robots, large blocks of elastomer for self-supporting and end-effectors lead to a comparatively low energy density (*12*); the widely used fluidic actuation also dissipates too much energy in viscous friction with the tube, especially when the actuation frequency is high (*13*). Three solutions are proposed to solve the speed and strength problem: larger energy input, higher frequency, and structural instability. For example, Shepherd et al. (*14*), Bartlett et al. (*15*), Tolley et al. (*16*), Keithly et al. (*17*), and Aubin et al. (*18*) used explosive actuators to power jumping robots. These robots usually move fast but in a very uncontrollable pattern. In addition, the soft materials accommodating the explosive impact degenerate rapidly after a few repetitions. Mosadegh et al. (*19*), Huang et al. (*20*), Li et al. (*21*), and Wu et al. (*22*) designed soft robots with high actuation speed or frequencies, but the highest achievable speed or frequency is limited by the moduli and elastic wave velocities of elastomers. Tang et al. (*23*) built a galloping soft robot with a high speed of 2.68 body length per second (BL/s) and a soft swimmer of 0.78 BL/s by utilizing instability. Some other research (*13*, *24–26*) also points to the potential of bi- and multistable mechanisms. Recently, Chi et al. (*24*) claimed a record-breaking speed of 3.74 BL/s (or 85.27 mm/s) for a tethered soft robotic fish with bi-stable mechanisms. However, this speed measurement can be mistaken because of a wrongly assumed body length (should be ~15 cm instead of the assumed 2 cm). A corrected speed would be 0.57 BL/s. With similar inspiration, our work shows a hair-clip-like bi-stable mechanism made by in-plane prestressed semi-rigid materials, which we term hair clip mechanisms (HCMs). They are promising to improve the dynamic (*1*) and static (*2*) performance of soft robotics (Fig. 1). The design method of HCM robots is proposed with the help of analytical solutions. A pneumatic soft fish robot and a untethered motor-driven one is fabricated using this method.

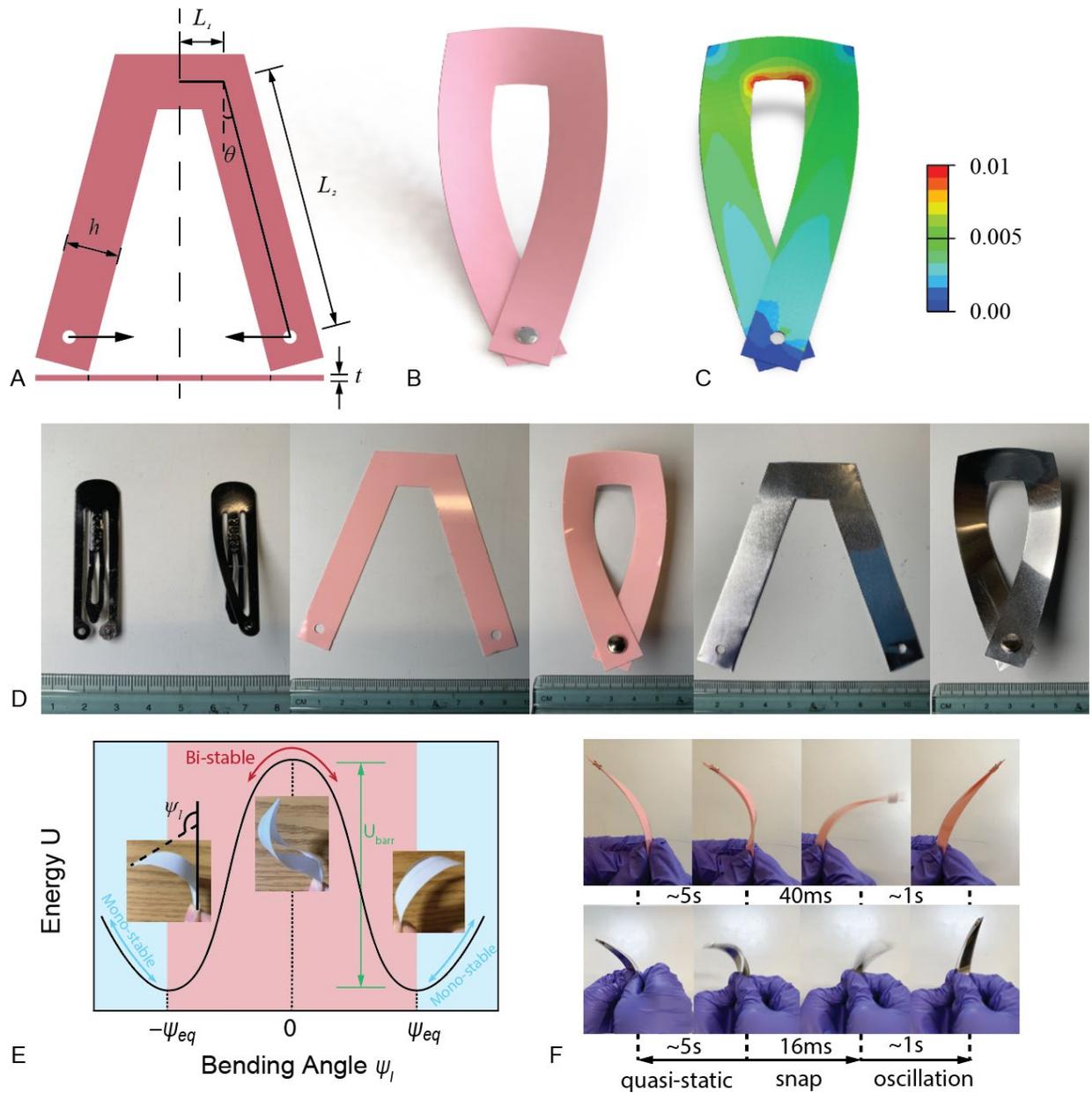

**Fig. 1 Design and principles of the bi-stable prestressed hair clip mechanism (HCM).** **(A)** The geometry of the ribbon before assembly. **(B)** and **(C)** Assembled configuration and corresponding contour of maximum principal strain, respectively. Dimensionless shape factors $\theta = 10°$ and $\gamma_s = L_2 / L_1 = 6$; other parameters $h/L_1$=15mm/12.5mm and $t/L_1 = 0.381$mm/12.5mm. **(D)** A commercial hair clip as well as plastic and steel hair clips before and after assembly ($\theta = 20°$ and $\gamma_s = 6$). **(E)** The energy profile and evolution of configuration of the snapping-through of a paper HCM. **(F)** The dynamics of plastic and metal HCMs' snapping under a high-speed camera ($\theta = 20°$ and $\gamma_s = 6$).

## Using HCM's bistability for repetitive propulsion

Enlightened by buckling and post-buckling knowledge (27, 28) and the fact that most fast-moving animals in nature need a hard skeleton, we demonstrate in Fig. 1 and movie S1 the geometry and assembly of a hair-clip-like in-plane prestressed mechanism that is fabricated with tensile-resistant materials like carbon fiber/epoxy, metal, plastic, and paper sheets. Even though these materials are high in modulus, the out-of-plane stiffness of their slender ribbons is usually smaller than the membrane stiffness by an order of $10^5\sim10^6$, making the ribbons bendable and compliant. The kinked ribbon deforms due to lateral-torsional buckling when the two extremities are closed and pinned together. Eventually, it becomes a bi-stable mechanism (Fig. 1D and movie S1). Two major assumptions are used in deriving the mathematical equations of HCM in the Supplementary Information (SI): small deflection (29) and kinked-to-straight assumption (fig. S1A). Both simplify the problem and are necessary for analytic solutions despite the errors they bring. According to the derivation in SI, the out-of-plane bending angle $\psi_l$ (Fig. 1E) of the ribbon tip ($z = l = L_1 + L_2$) can be calculated from

$$\psi_l \approx \left.\frac{du}{dz}\right|_{z=l} = -\frac{P_{cr}}{EI_\eta}\int_0^l \varphi(l-z)\,dz, \qquad (1)$$

with $P_{cr}$ being the critical load of the lateral-torsional buckling of the angled ribbon (SI), $EI_\eta$ being the bending stiffness along the $\eta$ axis, and $\varphi$ being the angular displacement of the section. The above equation decides the swing amplitude of the HCM during a snap-through. One indication of Eq. (1) is that $\psi_l$, a key factor for the kinematic performance of the HCM, is independent of the material modulus $E$ and dimensions like $L_1$, $L_2$, $h$, and $t$, but is only affected by the unitless shape factors $\theta$ (Fig. 1A) and $\gamma_s = L_2 / L_1$, giving space to the up-scaling and down-scaling of this mechanism. A detailed derivation and corresponding FEM verification is given in the SI. The estimation of Eq. (1) gives an error of around 5% (Fig. 2A and 2B) compared to the experimental measurements.

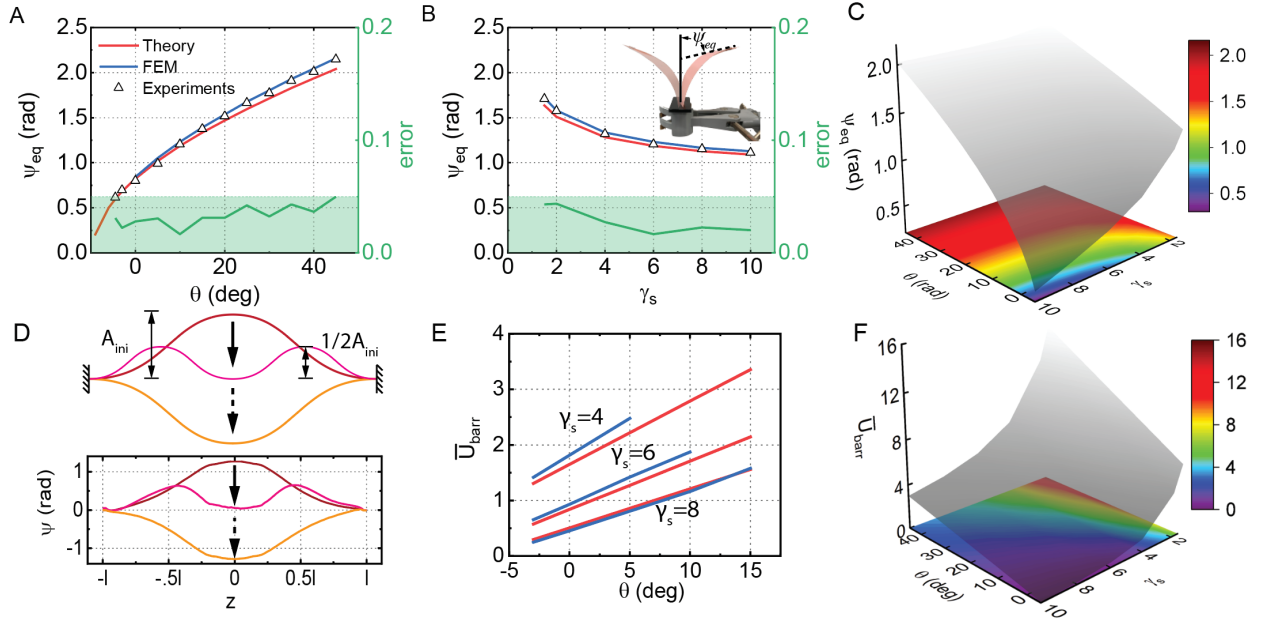

**Fig. 2 Characterization of the pre-buckling and the post-buckling of HCMs.** **(A)** and **(B)** Comparison of the value $\psi_l$ among theoretic model, FE reproduction, and experiment results w.r.t. varying prop angle $\theta$ when $\gamma_s = 6$ and dimensionless prop length $\gamma_s$ when $\theta = 10°$, respectively. The region within a 5% error (calculated by comparing the theory to experiments) is highlighted in green. The inset shows the bi-states and their tip bending angle $\psi_{eq}$. **(C)** and **(F)** Maps illustrating the designing spaces of $\theta$ and $\gamma_s$, respectively, w.r.t. $\psi_l$ and the unitless energy gap $\overline{U}_{barr}$. **(D)** Comparison of snap-through evolutions between the theory and the FE reproduction. $A_{ini}$ denotes the deflection "amplitude" of a pre-buckled curve. **(E)** Comparison between theory and FE results of $\overline{U}_{barr}$ w.r.t. shape factors $\theta$ and $\gamma_s$.

The energy barrier $U_{barr}$ between the bi-states (Fig. 1E) can be approximately calculated as

$$U_{barr} = U\big|_{\psi_l=0} - U\big|_{\psi_l=-\psi_{eq}\, or\, \psi_{eq}}$$
$$\approx 3U_{\psi_l=-\psi_{eq}\, or\, \psi_{eq}} \qquad (2)$$
$$= 3P_{cr} \cdot L_2 \left( \sin^{-1}\frac{1}{\gamma_s} + \theta \right),$$

assuming the configuration evolution of the buckling ribbon is like the snap-through process of a straight beam fixed at both ends (Fig. 2D). The comparison between the assumed and the simulated configuration change is shown in Fig. 2D. By defining the energy barrier to be the difference between the maximum and the minimum strain energy when HCM snaps, we can also calculate the energy barrier with FEM simulation and compare it to Eq. (2) (Fig. 2E), which gives a maximum error of approximately 10%. Again, the unitless energy gap $\overline{U}_{barr} = U_{barr} L_1 / EI_\eta$ also depends merely on $\theta$ and $\gamma_s$. For example, increasing the thickness $t$ by twofold doesn't change $\psi_l$ and $\overline{U}_{barr}$, and leads to an eight-fold increase in $U_{barr}$.

The time scale of the HCM snap-through is given by (*30*)

$$t_* \approx \frac{(2l)^2}{t\sqrt{E/\rho_s}} \qquad (3)$$

where $\rho_s$ is the density of the material. The difference in timescales of HCM snapping is shown with different materials in Fig. 1F and movie S2. For example, a plastic HCM with $\rho_s = 1.2$ g/cm$^3$, thickness $t = 0.381$mm, $E = 1.73$ GPa (SI), and total ribbon length $2l = 175$ mm takes about 40 ms to snap and the angular speed is $\sim 4 \times 10^3$ °/s. In comparison, a steel HCM with $\rho_s = 7.85$ g/cm$^3$, $t = 0.254$ mm, $E$ about 200 GPa, and $2l = 175$mm only takes about 16 ms and has an angular speed of $\sim 14 \times 10^3$ °/s (Fig. 1F). The corresponding theoretical answers from Eq. (3) are 67 ms and 24 ms, respectively, which are slightly larger than the experiments. These velocities are comparable to those of the throws of a professional baseball player of $\sim 9000$ °/s (*31*) and much faster than the tail beats of fish, which is about $100 \sim 1000$ °/s (*32*). In Fig. 1F, the tip bending angle $|\psi_l|$ of an HCM increases during the quasi-static energy-storing stage that takes ~5s. When the HCM snaps, the elastic energy is released in several to tens of milliseconds, generating fast reverse swinging and small shock waves that result in a clear snap sound (movie S2). Excessive energy is then dissipated through oscillation.

We note that the global buckling (snap-through) of HCMs can be triggered by a point displacement at the "core end" denoted by $L_1$ (Fig. 1A), which is convenient for their use as end effectors (*1*, *2*). HCMs accumulate elastic energy when they are actuated and snap rapidly when the actuation (pressure/displacement) exceeds a certain level, generating high speed at the "far end" denoted by $L_2$ (Fig. 1A). This local-global actuation is termed "bend-propagating actuation" in the relevant literature (*26*). HCMs can be possibly used in domains like origami/kirigami structures (*33*, *34*), deployable devices (*27*, *35*), morphing airfoils (*4*, *36*), etc.

## Swimming robots with HCMs

### Tethered pneumatic fish robot

Inspired by the resemblance of HCM snapping and fish undulation, we use HCMs for fish robots to demonstrate their function in aquatic propulsion. Since one of the most popular actuation methods for soft robotics is pneumatic actuation (*11*), we designed and tested in this section (Fig. 3) a pressure-driven HCM fish with dimensions of length × width × height = 18.6 × 6.5 × 5.2 cm$^3$ and a weight of 42.5 g. It comprises an HCM body (pink, film thickness $t_1$=0.381mm), a riveted fishtail (grey, film thickness $t_2$ = 0.191mm), and a hollow 3D-printed fish head with ballast. The HCM used in the fish has a geometry of $2l$ = 175 mm, $\theta = -3°$, and $\gamma_s = 6$, which yields a tip bending angle of $\psi_l = 39°$ and snapping time of ~68 ms from the theory. A pair of pneumatic bending actuators is attached to the HCM body antagonistically. When gas is pumped into these actuators, their bending deformation will snap the mechanism. An alternate actuation pattern of the actuators would enable the HCM fish to undulate. A detailed fabrication process of the pneumatic actuators is illustrated in the SI.

With a pressure of 150 kPa and a frequency of 1.3 Hz (period = 760 ms), the pneumatic HCM fish swims underwater at an averaged horizontal speed of 1.40 BL/s or 26.54 cm/s (calculated from photo sequence) which is twice as fast as the mono-stable reference swimmer with the same weight (~41.6 g) and same actuation conditions (0.69 BL/s or 13.10 cm/s, Fig. 3B and movie S3). The reference fish use a mono-stable plastic sheet made of the same material and actuated by the same actuators. Figure 3C depicts the peduncle swinging kinematics of the two fish robots, measured from high-fps videos (movie S4). A sinusoidal/triangular swinging pattern is observed in the reference swimmer, while the HCM fish shows a unique pattern that is different from the old ones (*37*). It is noted from the videos that the new swinging pattern gains more kinetic energy from each flapping than the reference, probably because water drag is proportional to the squared value of the flapping speed, i.e., $F_D \propto v^2$. The two fish robots have similar undulation amplitude (marked as red and blue in Fig. 3C), but their angular velocities diverge. The HCM fish flaps its tail at ~1200 °/s, about three times the speed of the reference one of 340 °/s.

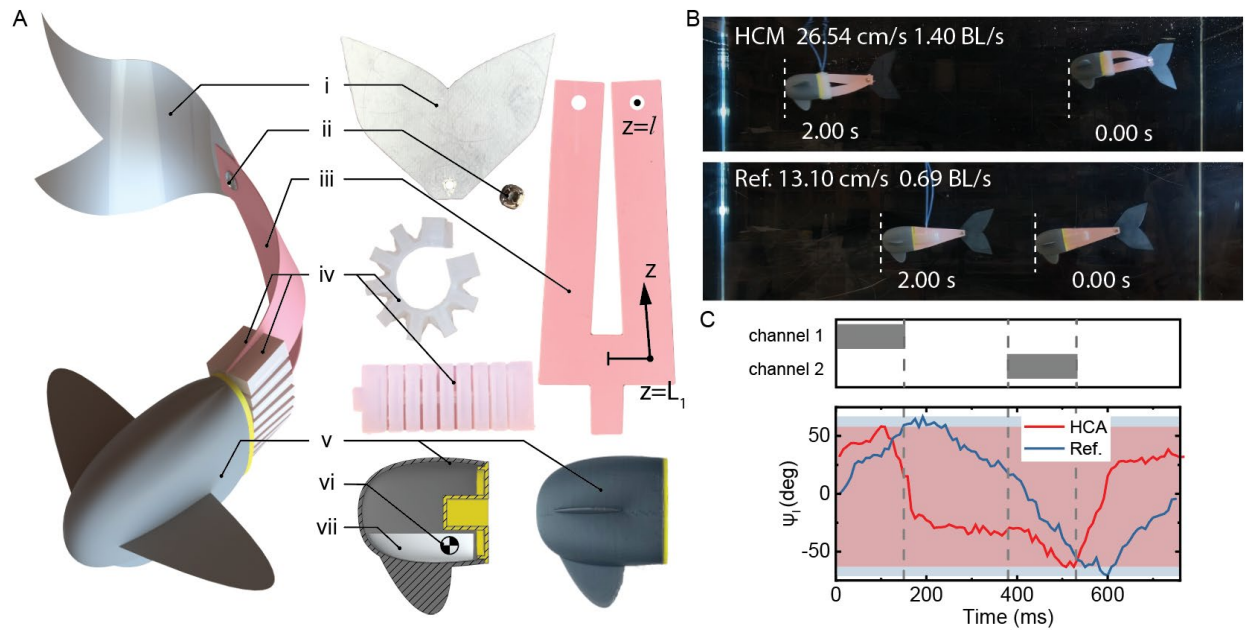

**Fig. 3 Pneumatic HCM fish robot and its comparison with the reference model. (A)** Assembly schematics of the robotic fish. i. caudal fin ($t = 0.191$ mm), ii. rivet pin, iii. HCM plate ($2l = 175$ mm, $t = 0.381$ mm, $\theta = -3°$, and $\gamma_s = 6$), iv. antagonistic pair of pneumatic bending units (fabrication and dimension details shown in methods and fig. S4), v. 3D-printed hollow fish head, vi. location of the mass center after assembly, and vii. cast ballast. **(B)** Comparison of underwater locomotion between HCM-based bistable fish robot and its mono-stable counterpart. Velocities are calculated from simple trigonometry. Scale bar, 150mm. **(C)** Comparison between the angular displacement $\psi_l$ patterns of the two prototypes under 150kPa and 1.3Hz pressurization (period = 760 ms). Grey areas show when a channel is pressurized. The red and blue areas show the angular displacement range of the HCM and reference model, respectively.

The energy barrier and the undulation frequency of the HCM decide the actuation pressure of the HCM fish robots. For example, the critical pressure needed to snap the HCM in a quasi-static situation is shown in fig. S7, where the increasing shape factors $\theta$ lead to a higher energy barrier and, thus, critical pressure. When the undulation frequency increases from 1.3 Hz (Fig. 3) to 2.5 Hz (movie S5), the actuation pressure required becomes 300 kPa, which is two times the pressure in the 1.3Hz undulation. The simulated processes of the assembly and the actuation of HCM are shown in movie S6.

Untethered motor-driven fish robot

Following the untethering trend of soft robots, we build a motor-driven fish robot that is length × width × height = 21.5 × 4.5 × 12.0 cm³ in size and 125 g in weight (Fig. 4A and fig. S5). Similarly, the robot is made of three parts: an HCM body (coral colored, film thickness $t_1$ = 0.762 mm), a riveted fishtail (blue, film thickness $t_2$ = 0.127 mm), and a 3D-printed fish head. A 7.4 V battery set, a BLE microcontroller, and a servo motor are included in the hollow head to make the robot self-contained. The HCM used in the fish has a shape of $2l$ = 174 mm, $\gamma_s$= 2, and $\theta$ = −23.5°, which yields a tip undulation amplitude of $2\psi_l$ = 68° and a snapping time of 34 ms from Eq. (1) and (3). The horn of the onboard mini-servo (part v in Fig. 4A) passes through the opening of the waterproof layer (part vii in Fig. 4A) made of silicone rubber. The core area of the HCM body is actuated by the servo towards left and right alternately so that the fish body and fin swing to right and left accordingly (Fig. 4B, 4C, and movie S7).

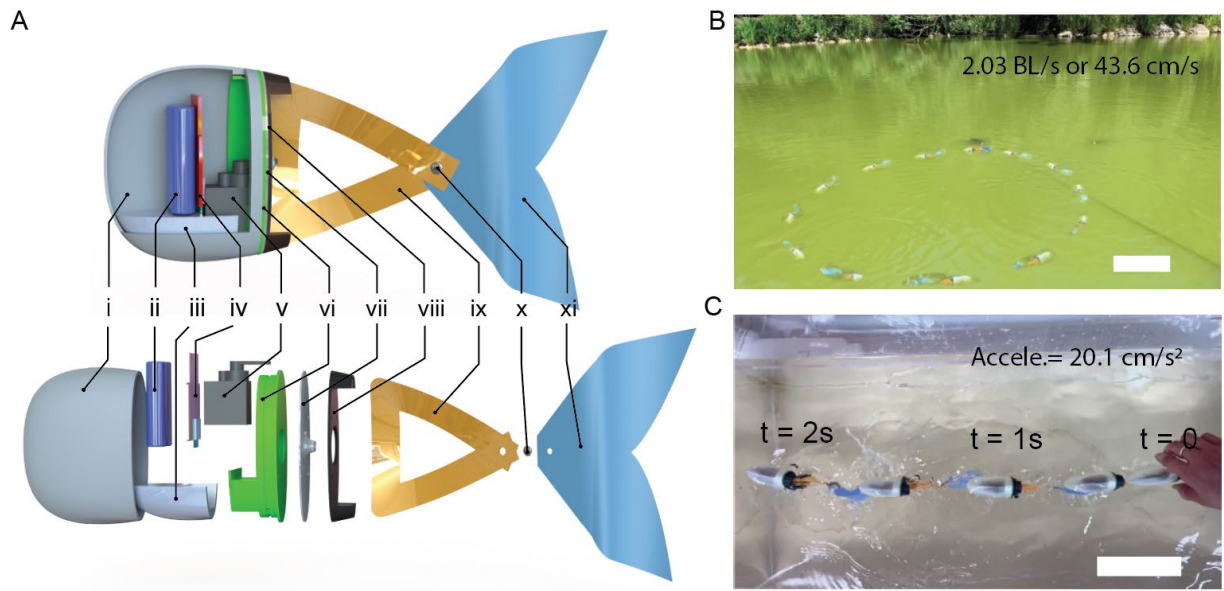

**Fig. 4 The assembly and operation of the untethered fish robot. (A)** Assembly schematics of the robot. i. hollow fish head, ii. customized battery pack (7.4 V, 350 mAh, 2-in-1 pack), iii. cast ballast, iv. BLE microcontroller, v. mini servo, vi. servo holder, vii. waterproof layer, viii. connecting layer, ix. HCM plate ($t$ = 0.762, $\theta$ = − 23.5°, $\gamma_s$ = 2, and $\psi_l$ = 34°), x. steel rivet, and xi. caudal fin ($t$ = 0.127 mm). Photos of the robot are given in the Supplementary Information (SI). **(B)** The circular cruising sequence of the robot at an undulation frequency of 3Hz and a speed of 2.03 BL/s or 43.6 cm/s. **(C)** The straight swimming sequence of the robot in an aquarium tank at an undulation frequency of 3Hz and an average acceleration of 20.1 cm/s². Due to the different resistance, the buckling timescales are t* ≈ 17 ms in the air and t* ≈ 50 ms in the water. Scale bars = 5 cm.

The fish can swim underwater, but to remotely control it with a BLE device, we make its density smaller than the water so it can float close to the surface. The circular swimming path of the fish in Fig. 4B is for better filming and better speed measuring from the frame sequence. With a tail beat of 3Hz (6 times of snapping per period), it takes the robot 12.7s to cover the circular path of 554 cm, which corresponds to an overall speed of 2.03 BL/s or 43.6 cm/s, and the instantaneous speed and average speed are given in fig. S5C. The straight swimming of the fish robot is shown in Fig. 4C and move S8, with a starting acceleration of about 20.1 cm/s$^2$. Interestingly, our fish robot doesn't scare away local creatures, which can benefit its non-invasive observation of aquatic life. More undulation movies of the robot are given in movie S9.

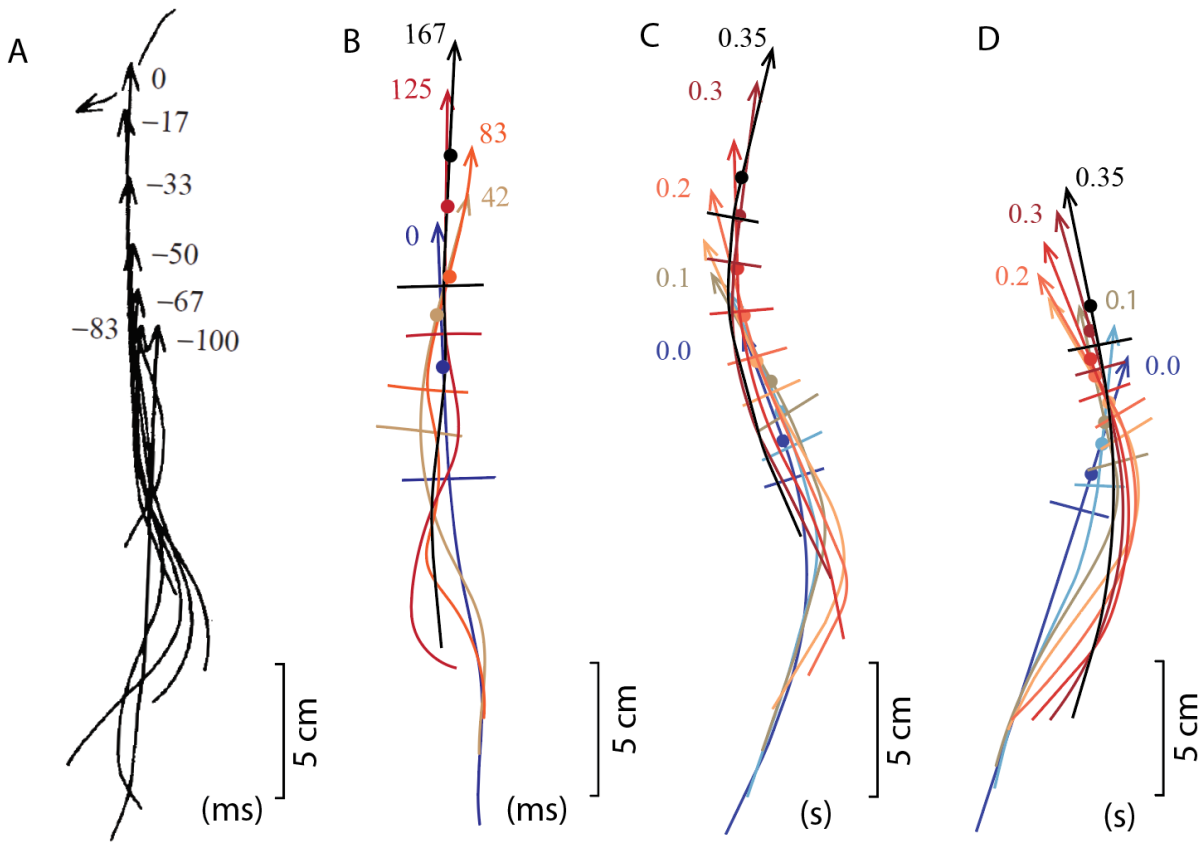

**Fig. 5 Comparison of strike patterns between biological and robotic fish.** (A) Configuration sequence of a pike (*Esox* sp.) during its hunting with a single strike. Times are in milliseconds. Reproduced from Webb and Skadsen (*38*). (B) Configuration sequence of the untethered fish robot with an undulation of 3Hz. Solid dots denote the centers of mass, and short bars mark the connection between fish heads and bodies. Times are in milliseconds. (C) and (D) Configuration sequences of the pneumatic HCM fish and the reference fish undulating at 1.3Hz, respectively. Times are in seconds.

Comparisons of patterns and velocities

The previous experiments show that the thrust-to-velocity-squared rule makes this bi-stable undulation better than sinusoidal ones for obtaining propulsion. Figure 5 juxtaposes the swimming patterns of a real fish and the three imitators. Although it is well accepted that the sinusoidal undulation is the most energy-saving cruising pattern for Body and/or Caudal Fin (BCF) swimmers (*37*, *39–42*), fish also use strikes with high power rates to generate large thrust and speed in behaviors like hunting and escaping (*38*, *43*, *44*). For example, Fig. 5A is the configuration sequence of a pike (*Esox* sp.) during its single-strike hunting, and the 3Hz bi-stable undulation pattern in Fig. 5B resembles the quick start of the pike. While the sinusoidal wave is preferred by nature in cruising, human inventions may have a different optimal solution because they don't have the living-body constraints that organisms are subject to (*45*). The bi-stable undulation makes the untethered fish robot comparable to biological fish in speed (2~10 BL/s) (*46*); thus, it can be a novel strategy for efficient bionic underwater vehicles, especially when the benefit of simplicity of the HCM is considered. However, the patterns shown in Fig. 5C and 5D diverge from real fish undulation due to low actuation frequency and large recoil of the fish head, respectively.

A comparison of speed and frequency is shown in Fig. 6. The hollow circles in the figure are for tethered swimmers, and the solid circles are for untethered ones. Contrary to the notion that untethering a soft robotic swimmer decreases its swimming capacity (*21*), in our case, the untethered motor-driven soft fish has a higher speed than the tethered one because the servo is more efficient than a pneumatic power source. Our untethered soft swimmer presents a faster speed than previous counterparts, to the best of the authors' knowledge, outperforming the previous ones by threefold (294%) and twofold (186%) in BL/s (*21*) and cm/s (*46*), respectively, but is still slower than tethered soft swimmers (*47*). A similar study presented by Chi et al. (*24*) claimed a speed of 3.74 BL/s by mistakenly using the "width" of the robot as the robot's body length, which resulted in an overestimated value. A corrected speed would be 0.57 BL/s, shown as a blue circle in Fig.6. The major difference between their work and ours includes actuation methods, mathematical modeling, designing principles, etc., which lead to differences in performance and applications.

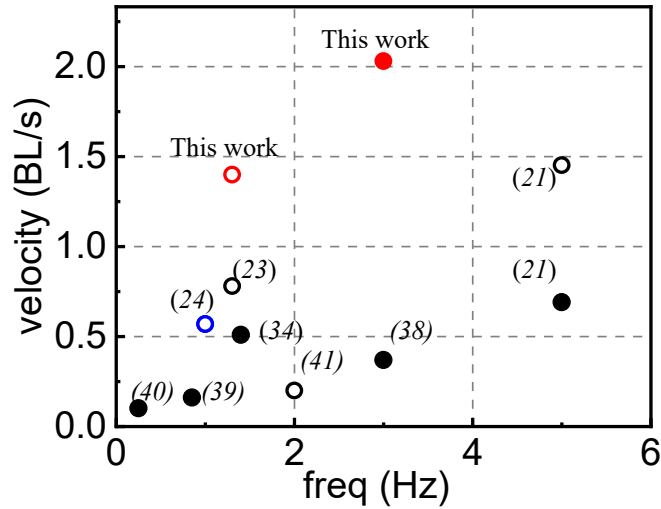

**Fig. 6 Comparison of swimming velocity w.r.t actuation frequency among the HCA-driven and various reported soft robotic swimmers.** The hollow circles in the figure denote tethered swimmers, and solid circles denote untethered swimmers. The red-colored symbols are fishes in this work. The blue symbol is the corrected speed of (*24*).

## Conclusion

Harnessing the prestressing and bi-stability of 2D materials opens a new way to design mechanisms with enhanced capability and functionality and can inspire applications in diverse fields and scales. Our research provides principles for designing and fabricating a hair-clip-inspired mechanism, which we term HCM, and demonstrates the advantages of HCMs as the propulsion method for fish robots. The bi-stable HCM propeller provides the pneumatic fish robot with a two-fold increase in cruising speed. The untethered soft swimmer with an HCM body swims more than three times faster than previous studies. The in-plane prestressed HCM has the advantages of structural rigidity (*1*), motion capacity, and design simplicity. The HCM undulation can inspire novel swimming patterns for underwater vehicles, which can be a breach for soft swimmers to compete with real fish and participate in the revolution of soft robotics.

## Acknowledgments

We especially thank Yichao Tang for their help with the experiments, Jiefeng Sun for the simulation suggestion, and all colleagues for the inspiring discussions. **Funding**: This work was supported by the Earth Engineering Center and the Center for Advanced Materials for Energy and Environment at Columbia University. **Author contributions:** Z. X. conceived the presented idea, developed the theory, built the simulation, and wrote the manuscript. Z. X. and L. C. designed and carried out the experiments. H. L. supervised the project. **Competing interests:** There are no conflicts of interest to declare. **Data and materials availability:** All data are available in the manuscript or the supplementary materials.